\documentclass[isre,nonblindrev]{informs3}

\DoubleSpacedXI

\usepackage{natbib}

 \bibpunct[, ]{(}{)}{,}{a}{}{,}

 \def\bibfont{\small}

 \def\bibsep{\smallskipamount}

\TheoremsNumberedThrough

\EquationsNumberedThrough

\MANUSCRIPTNO{1}

\usepackage{tabularx}
\usepackage{threeparttable}
\usepackage[table,xcdraw]{xcolor}
\usepackage{colortbl}
\usepackage{float}
\usepackage[colorlinks=true, linkcolor=black, citecolor=black, urlcolor=black]{hyperref}
\usepackage{booktabs}
\usepackage{tcolorbox}
\usepackage{enumitem}

\tcbuselibrary{most}

\usepackage{dsfont}
\usepackage{setspace}

\usepackage{amsfonts, comment}

\tcbuselibrary{most}

\begin{document}

\RUNAUTHOR{Cheng, Wang, and Ghose}

\RUNTITLE{LLMs for Explainable Business Decision-Making}
\TITLE{LLMs for Explainable Business Decision-Making: A Reinforcement Learning Fine-Tuning Approach}

\ARTICLEAUTHORS{
\AUTHOR{Xiang Cheng, Wen Wang}
\AFF{Robert H. Smith School of Business, University of Maryland\\ \EMAIL\{xccheng, wenw\}@umd.edu \URL{}}

\AUTHOR{Anindya Ghose}
\AFF{Leonard N. Stern School of Business, New York University\\ \EMAIL
aghose@stern.nyu.edu \URL{}}
}

\ABSTRACT{
\looseness -1 \looseness -1
Artificial Intelligence (AI) models increasingly drive high-stakes consumer interactions, such as credit approvals, yet their decision logic often remains opaque. Prevailing explainable AI techniques rely on post hoc numerical feature attributions, which fail to provide coherent narratives behind model decisions. Large language models (LLMs) present an opportunity to generate natural-language explanations, but three design challenges remain unresolved: explanations must be both decision-correct and faithful to the factors that drive the prediction; they should be able to serve multiple audiences without shifting the underlying decision rule; and they should be trained in a label-efficient way that does not depend on large corpora of human-scored explanations.
To address these challenges, we introduce LEXMA (LLM-based EXplanations for Multi-Audience decisions), a reinforcement-learning-based fine-tuning framework that produces narrative-driven, audience-appropriate explanations. LEXMA combines reflection-augmented supervised fine-tuning with two stages of Group Relative Policy Optimization (GRPO). Specifically, it fine-tunes two separate parameter sets to improve decision correctness and satisfy stylistic requirements for different audiences, using reward signals that do not rely on human-annotated explanations.
We instantiate LEXMA in the context of mortgage approval decisions. Results demonstrate that LEXMA yields significant improvements in predictive performance compared with other LLM baselines. Moreover, human evaluations show that expert-facing explanations generated by our approach are more risk-focused, and consumer-facing explanations are clearer, more actionable, and more polite.
Our study contributes a cost-efficient, systematic LLM fine-tuning approach to enhance explanation quality for business decisions, offering strong potential for scalable deployment of transparent AI systems.
}

 \thispagestyle{empty}

\maketitle

\newpage \setcounter{page}{1}

\section{Introduction}

Artificial intelligence (AI) systems increasingly drive high-stakes marketing decisions, including targeting, credit and eligibility assessment, and dynamic pricing \citep{davenport_how_2020, huang_strategic_2021, puntoni_consumers_2021}.
In these settings, a lack of transparency in model behavior can have serious consequences for firms and their stakeholders, such as regulatory penalties, perceived unfairness, and reduced consumer trust \citep{rudin_stop_2019}.
Consequently, explainable AI, which seeks to make model decision mechanisms understandable to humans \citep{martens_tell_2025}, has become a central concern for organizations and regulators \citep{bauer_explained_2023, mohammadi_regulating_2025}.

Most established techniques in explainable AI focus on producing numerical feature attributions. For instance, the Shapley additive explanations (SHAP) value method assigns weights to features according to their contribution to model predictions \citep{lundberg_unified_2017}.
Although widely used in practice (e.g., \citealp{senoner_using_2022}), these methods have several limitations.
For one, attribution methods are predominantly post hoc. They approximate a black-box model that has already been trained, rather than embedding explanatory logic inside the decision-making process itself.  This approximation inherently limits their explanatory power and reliability \citep{slack_fooling_2020}.
Moreover, most attribution techniques implicitly assume structured, tabular inputs with predefined features and therefore cannot naturally handle unstructured inputs such as free text.
Finally, most attribution-based explanations are numeric. They highlight which features are important but do not naturally produce coherent, narrative rationales.
In marketing settings, this lack of narrative explanation limits the usefulness of AI systems for both consumers and managers, whose perceptions of fairness and transparency shape trust, loyalty, and adoption of algorithmic decisions \citep{martens_tell_2025,rai_explainable_2020}.

Recent advances in generative AI, and in large language models (LLMs) in particular, create an opportunity to move beyond numeric explanations by generating natural-language rationales for complex decisions \citep{gat_faithful_2024,koa_learning_2024,wang_llm-gan_2024}.
However, three design challenges remain unresolved in using LLMs to explain decisions.  First, explanations must be both decision-correct and decision-aligned. Predictive accuracy is a prerequisite for any useful explanation, yet deep learning models, including LLM-based approaches, often underperform Gradient Boosting methods on tabular data \citep{grinsztajn_why_2022}. At the same time, an LLM that simply predicts and then generates a narrative about its prediction can produce fluent but unfaithful rationales that are not the real basis for its decision \citep{turpin_language_2023}. A practical system therefore needs an architecture and training scheme in which the model can achieve high predictive performance and in which the explanation is tightly coupled to the factors that drive the decision, rather than functioning as a post hoc justification.

Second, in many customer-facing contexts, decisions must be communicated to multiple audiences. The same credit decision may be read by loan officers and consumers, who require different levels of detail, terminology, and tone, yet the underlying decision rule must remain stable. Simply re-prompting an LLM with different roles, for example ``speak as an underwriter'' versus ``write to the applicant'', can change both the explanation and the decision, which compromises the robustness of the system \citep{bauer_explained_2023}.

Third, organizations need a way to improve explanation quality that is label-efficient. Using a large corpus of human-annotated explanations to fine-tune a model is costly; a scalable approach should therefore rely primarily on existing decision labels rather than extensive human-labeled explanation annotations.
Motivated by these challenges, we ask the following research question: \textit{How can we fine-tune large language models in a cost-efficient way so that they deliver decision-correct, decision-aligned explanations and support audience-specific communication?}

\begin{figure}[htb!]
  \centering
  \caption{Illustration of a Mortgage Decision Workflow and Our Goal}
  \includegraphics[width=0.8\linewidth]{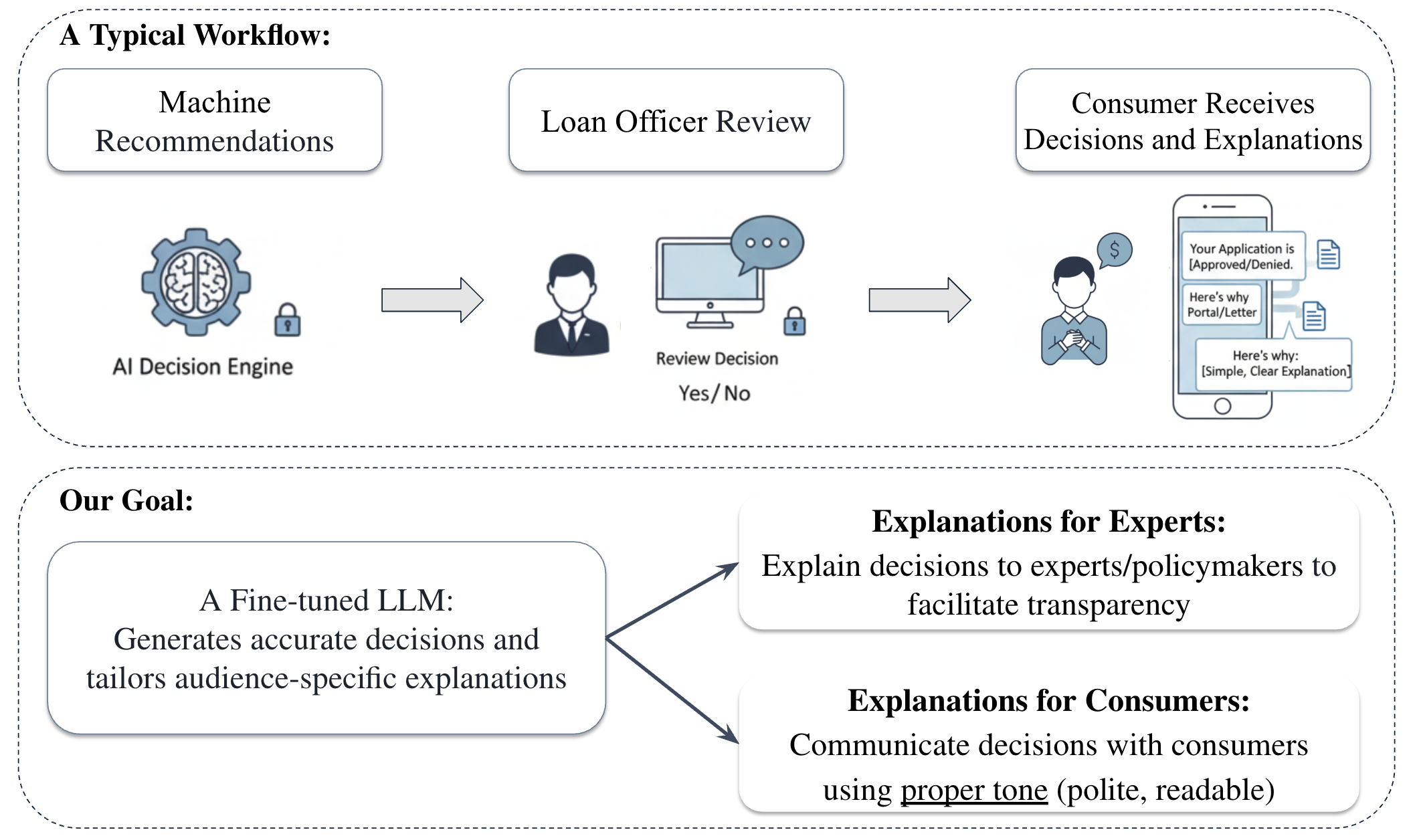}
  \label{fig:goal_illustration}
\end{figure}

To illustrate these challenges, consider mortgage approval as a concrete decision context.
In typical practice, an AI system screens applications and proposes a preliminary approve or deny recommendation. A loan officer reviews the AI recommendation and makes the final decision. The consumer then receives the decision and relevant explanations. The top panel of Figure \ref{fig:goal_illustration} depicts this workflow.
In this  context, each of the three challenges is salient. Decision-correct and decision-aligned explanations are critical because expert reviewers rely on the system's rationale to triage files. There are inherent multi-audience needs for decision communication: internal professionals require precise, risk-focused rationales that cite case-specific evidence, while consumers need plain, respectful, and actionable guidance that avoids jargon yet remains consistent with the same decision rule. Label-efficiency is also essential, since institutions rarely maintain large corpora of human-written explanations paired with quality scores. The bottom panel of Figure \ref{fig:goal_illustration} summarizes our objective: to fine-tune a large language model that generates accurate decisions and audience-specific explanations for experts and consumers.

To address these challenges, we design \textbf{LEXMA} (LLM-based EXplanations for Multi-Audience decisions), a multi-objective fine-tuning framework for LLM-based explainable decision-making. LEXMA addresses each of the three challenges above. To address the need for decision-correct and decision-aligned explanations, it treats the decision as the end of a joint reasoning and explanation process: the model first produces an internal reasoning trace, then generates a concise, audience-appropriate explanation, and finally issues a prediction that is conditioned on the reasoning and explanation. Training is implemented so that rewards for correct decisions are applied to the entire Reasoning $\rightarrow$ Explanation $\rightarrow$ Prediction trajectory, which encourages explanations that highlight case factors associated with accurate predictions and improves predictive performance.

To address the multi-audience requirement, LEXMA fine-tunes two separate parameter sets within the model, which we refer to as adapters. The two-adapter architecture includes a correctness adapter (ACC) and a tone adapter (TONE). The ACC adapter captures the decision boundary and risk-focused content and is shared across both expert-facing and consumer-facing explanations, while the TONE adapter adjusts communication styles for consumer-facing explanations. Expert-facing explanations are generated by  activating only ACC, whereas consumer explanations are generated by activating ACC and TONE together, so explanations differ in style and level of detail across audiences but the underlying decision rule remains consistent. To address label-efficiency during training, LEXMA combines reflection-augmented supervised fine-tuning with Group Relative Policy Optimization (GRPO) \citep{guo_deepseek-r1_2025}. Reflection-augmented supervised fine-tuning uses a strong reference model to provide structured targets without human-written explanations, and GRPO optimizes over groups of sampled trajectories using rewards that depend only on ground-truth decisions and simple rule-based tone metrics (for instance, readability and politeness). This avoids the need for human-scored explanation labels or learned reward models while still improving both decision accuracy and explanation quality.

We validate our approach within the context of loan approval decisions using a real-world dataset.
Loan approval is a particularly suitable high-stakes domain since AI systems are already extensively deployed \citep{fu_crowds_2021}, and approval decisions shape customer acquisition and long-term relationship value \citep{wei_credit_2016}.
Intuitively, our objective is to fine-tune an LLM to accurately predict loan approval outcomes and generate human-like explanations for both expert and consumer audiences. Here, a good explanation mirrors concise, context-sensitive rationales typically offered by loan officers.
Specifically, we utilize the publicly available HMDA dataset, a large-scale public dataset that includes rich structured financial data alongside approval or denial decisions from mortgage applications. We fine-tune the Qwen3-4B model \citep{yang_fine-tuning_2024}  under the LEXMA framework.

Our empirical findings confirm the effectiveness of the proposed framework. First, on decision correctness, our fine-tuned model delivers strong predictive performance; it achieves F1 scores of 0.897 under the expert prompt and 0.893 under the consumer prompt, surpassing raw Qwen model (F1 scores of 0.723, 0.726 under the expert and consumer prompts, respectively) and GPT-5 (F1 scores of 0.730, 0.771 under the expert and consumer prompts, respectively) and approaching strong tree prediction baselines.
Second, for expert explanations, the model produces concise, risk-focused rationales that practitioners prefer. We recruit loan professionals to assess the explanations generated by a raw Qwen model and our fine-tuned model. Experts consistently rate our model's explanations higher on risk relevance, decision appropriateness, and explainability, indicating that the narratives surface case-critical evidence that supports the chosen decision.
Third, for consumer explanations, the model communicates in plain, respectful, and actionable language. Human evaluation shows that consumers prefer our model's explanations over baseline explanations along various dimensions, including clarity, actionability, fairness, trustworthiness, reasonableness, and satisfaction. Tone tuning drives readability to plain-language levels and raises politeness while keeping the decision fixed.

Our study makes three main contributions to the methodological work in marketing and to the explainable AI literature. First, we introduce LEXMA as an AI artifact for high-stakes decision-making explanations. LEXMA is a systematic fine-tuning architecture that utilizes reinforcement learning and modular adapters to align decisions and audience-specific explanations in a label-efficient way. It moves beyond post hoc attribution tools toward an integrated, narrative-driven explainable system that is suitable for business contexts such as lending.
Second, we derive and validate a set of generalizable design principles for building label-efficient, multi-audience explanation systems with LLMs. These principles include: using comparison-based reinforcement learning with outcome labels to align explanations and decisions; and separating decision correctness from communication style through modular adapters, so that consumer-facing tone can be improved without shifting the underlying decision rule. These mechanisms provide prescriptive guidance for future AI artifacts that need to explain decisions to heterogeneous stakeholders under a single, stable decision policy.
Third, we provide empirical evidence on the performance and perceived usefulness of such an artifact in a real-world mortgage approval setting. We show that LEXMA approaches the predictive performance of strong tree-based baseline models such as XGBoost while outperforming foundation models such as GPT-5, and that expert underwriters and prospective consumers both prefer its explanations to those of an untuned LLM along multiple dimensions, including risk relevance, appropriateness, clarity, and trust.

Our paper proceeds as follows. Section \ref{sec:lit} reviews relevant literature and identifies literature gaps, Section \ref{sec:method} presents the proposed fine-tuning framework,  Section \ref{sec:eval_strategy} describes the applied dataset, experimental settings, and how we evaluate the results, and Section \ref{sec:eval_result} demonstrates results. Section \ref{sec:conclusion} concludes by discussing our results and directions for future research.

\section{Literature Review} \label{sec:lit}
\subsection{Explanations in AI Decisions}
Organizational decision-making that relies on AI requires explanations to support accountability, enhance trust, and enable effective human oversight \citep{bauer_mirror_2024}. A broad stream of behavioral research shows that users may resist or accept algorithmic advice depending on framing and perceived control \citep{bockstedt_humans_2025}, which makes the provision and form of explanations crucial for AI adoption \citep{lu_1_2025, wang_power_2025}.
Recent work in marketing documents that consumers react differently to algorithmic versus human decisions, and that perceived transparency and fairness strongly shape their responses in customer-facing decisions such as credit approval and platform access \citep{yalcin_thumbs_2022}.
Regulatory requirements also increasingly mandate explanations for certain decisions across domains; for example, adverse-action rules in U.S. consumer finance obligate lenders to provide specific reasons for denials.

A large literature in explainable AI has developed post hoc techniques that approximate model behavior. Feature-attribution methods such as LIME and SHAP assign importance scores to inputs near a focal prediction, while gradient-based approaches like Integrated Gradients aim to trace contributions through differentiable networks \citep{lundberg_unified_2017}. A complementary line of research constructs contrastive and counterfactual explanations that identify minimal changes needed to flip a decision \citep{wachter_counterfactual_2018}.

Despite the broad adoption of these approaches (e.g., \citealp{senoner_using_2022}), they face well-documented limitations. First, attribution and saliency techniques can be brittle to perturbations, sensitive to model randomization, and vulnerable to adversarial manipulation, which raises concerns about faithfulness and robustness of the model explanations \citep{rudin_stop_2019}. Moreover, many approaches presuppose tabular inputs with clearly defined features, which complicates their use when decisions depend on unstructured evidence such as free text \citep{netzer_when_2019}. Most importantly, these methods serve the objective of facilitating  understanding of certain decisions made by the AI system, yet numeric attributions often fail to supply coherent, audience-specific narratives that connect case facts to the decision rule in ways that a broad audience can understand and act on \citep{martens_tell_2025}. These limitations indicate a gap between the prevailing use of post hoc numeric rationales and the demand for faithful, case-grounded narrative explanations that meet the diverse needs of organizational audiences. We address this gap by embedding explanation into the decision-generation process and by using LLMs to produce concise, audience-tailored narratives rather than relying exclusively on post hoc feature scores.

\subsection{Large Language Models for Explaining Decisions}
Recent advances in LLMs show promise for providing explanations, as these models can generate fluent natural language and perform reasoning tasks. Unlike numeric-based explanation methods, LLMs can articulate why a decision was made in a human-readable way, drawing on broad knowledge and language generation abilities \citep{li_frontiers_2024}. Techniques such as chain-of-thought prompting elicit stepwise reasoning that can improve problem solving and provide a foundation for generating explanations \citep{wei_chain--thought_2023}. Building on these capabilities, one line of work uses pretrained LLMs directly and integrates them with architectural scaffolds to enhance explanation quality. For instance, in the context of misinformation detection, LLMs have been embedded in adversarial pipelines that generate candidate content, critique it, and in the process highlight decision-relevant information while providing textual justifications \citep{wang_llm-gan_2024}. Other studies prompt LLMs to generate counterfactual variants of inputs and then examine how predictions change, yielding contrastive, user-facing explanations that identify minimal changes relevant to the decision \citep{gat_faithful_2024}.

Despite these opportunities, two issues limit LLM-based explanations in practice. First, faithfulness and traceability remain difficult: the model itself is a black box \citep{freedman_argumentative_2025}, and it can produce convincing explanations that do not correspond to its actual decision process \citep{turpin_language_2023}. Ensuring that an explanation is not merely plausible but faithful to the decision procedure is challenging. Second, generic LLMs may underperform on certain decision tasks without specialization. Tabular financial data, such as information in loan applications, often poses a challenge to generic LLMs \citep{shwartz-ziv_tabular_2022}. For many tabular datasets, even sophisticated deep learning models struggle to outperform Gradient Boosting decision trees \citep{grinsztajn_why_2022}. In high-stakes settings, predictive accuracy is a prerequisite for useful explanations; an elegant explanation is useless if the prediction is wrong.

In response to these challenges, a growing stream of work fine-tunes LLMs for specific decision and explanation tasks. Early work uses supervised fine-tuning to teach models, using explanation pairs, to summarize case evidence and articulate reasons, drawing on human-annotated rationale corpora \citep{camburu_e-snli_2018}.
Because human annotation of reasoning is expensive, especially in specialized domains such as lending, recent research relies on reinforcement learning to enhance explanation quality, and explores approaches that do not require human annotation. For example, \citet{koa_learning_2024} train an LLM to predict stock movements while generating explanations. The model is prompted to summarize key information, generate both explanations and decisions, and then the model is fine-tuned using PPO (Proximal Policy Optimization).
Even with these advances, the extant literature largely overlooks audience adaptation in explanations. This is crucial since the usefulness of explanations depends on audience needs and communicative style; for instance, laypeople and domain experts often require different levels of detail, terminology, and tone, so a single explanation format is unlikely to be effective \citep{bauer_explained_2023}.

In summary, these observations reveal a gap in the current literature: there is no systematic fine-tuning approach that simultaneously ensures competitive predictive performance, label-efficient training of explanations, and audience-appropriate communication while safeguarding faithfulness. We address this gap by coupling decision making and explanation within a single generation process, using comparison-based reinforcement learning to improve correctness without explanation labels, and separating correctness from tone through modular adapters that support audience-specific narratives in high-stakes marketing decisions.

\section{Methodology} \label{sec:method}

\subsection{Overview of LEXMA}
The primary objective is to fine-tune a large language model to generate high-quality explanations for mortgage approval decisions. High-quality explanations must be factually grounded and appropriate for the target audience. A necessary condition for this objective is that the underlying decision is correct. Decision correctness is therefore treated as a supporting objective that enables explanation quality.
We refer to this overall fine-tuning framework as \textbf{LEXMA} (LLM-based EXplanations for Multi-Audience decisions).

Training proceeds in three stages. In the first stage (SFT), we apply reflection-augmented supervised fine-tuning to anchor the model to the task and a fixed output schema. In the second stage (GRPO-Step1), we apply Group Relative Policy Optimization on a correctness adapter (ACC) under both expert and consumer prompts to improve decision accuracy while the base model remains frozen. In the third stage (GRPO-Step2), we apply GRPO on a tone adapter (TONE) with the ACC adapter active for generation but frozen for updates and with advantages calculated based on explanations, which improves consumer-facing style without shifting the decision boundary.

\begin{figure}[htb!]
  \centering
  \caption{The LEXMA Fine-tuning Framework}
  \includegraphics[width=0.9\linewidth]{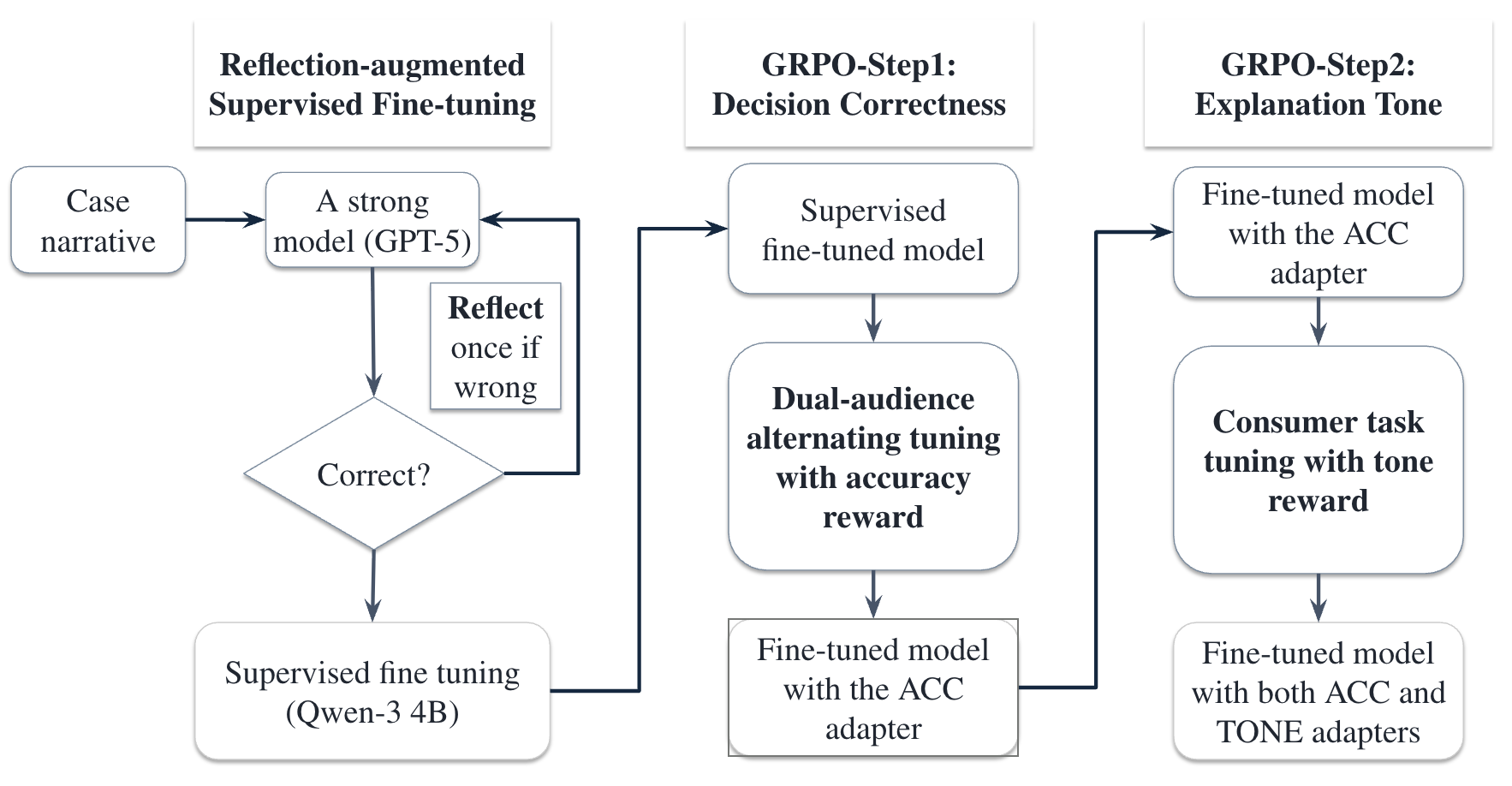}
  \begin{flushleft}
    \footnotesize \parbox{1.0\linewidth}{ \vspace{1em}
      \textit{Notes.}
      The LEXMA framework covers three stages: (1) reflection augmented SFT that creates a correctness backbone; (2) GRPO-Step1 that trains the ACC adapter for decision correctness using GRPO; (3) GRPO-Step2 that trains the TONE adapter for consumer explanation tone with the ACC adapter frozen. The prompt and output are always structured as Reasoning $\rightarrow$ Explanation $\rightarrow$ Prediction so that decisions are grounded in their narratives.
    }
  \end{flushleft}
  \label{fig:three_step_framework}
\end{figure}

\subsection{Problem Setup}
We first formally present the problem.
Let the dataset be
\(\mathcal{D} \,=\, \bigl\{(x_i, y_i)\bigr\},\) where $i$ denotes a data entry.
Consider the context of mortgage applications. Each \(x_i\) aggregates structured attributes (for example credit score, loan-to-value, debt-to-income) and unstructured text from the application file, while \(y_i\in\{\text{Approve},\text{Deny}\}\) indicates the ground-truth decision. For each case the model must emit a concise, audience-aware explanation \(e_i\) and a decision \(\hat y_i\in\{\text{Approve},\text{Deny}\}\).  We map Approve to 1 and Deny to 0 when computing rewards.
We use \(\pi_\theta\) to denote the policy of the model.

We introduce a three-phase generation schema that we consistently use as an output formatting requirement during fine-tuning. This schema can be summarized as follows: for each case \(i\) with inputs \(x_i\), the model first samples an internal reasoning chain \(c_i\) that breaks the task into intermediate steps, then writes an explanation \(e_i\), and finally issues a decision \(\hat y_i\).
We next describe these phases in detail.

In the first phase (Reasoning), we exploit the inherent chain-of-thought capability of a reasoning model.  For each \(x_i\), the model produces an internal reasoning trace
\[
c_i \sim \pi_\theta\bigl(\,\cdot\mid x_i\bigr),
\]
which decomposes a complex financial decision into intermediate substeps.  Prior work has shown that such internal CoT significantly improves multi-step reasoning and overall task performance \citep{wei_chain--thought_2023}.  However, these traces are overly detailed and include many redundant or low-salience tokens, rendering them unsuitable as end-user explanations \citep{yang_fine-tuning_2024}.

In the second phase (Explanation), we instruct the model to distill its hidden reasoning trace into a concise narrative.  Conditioned on \(x_i\) and \(c_i\), it generates
\[
e_i \sim \pi_\theta\bigl(\,\cdot\mid x_i, c_i\bigr).
\]
Each explanation \(e_i\) highlights only the most critical risk factors on a case-by-case basis in language reminiscent of an experienced loan officer’s rationale.
Predefining a set of features that should be included in these explanations is impractical, since there may be too many features and only some of them reveal important information for the specific loan case. Therefore, which features should be mentioned in the explanation is learned by the model through the fine-tuning process.
By focusing on key features and enforcing brevity, the model learns to omit repetition and irrelevant details, producing explanations that are both informative and succinct.

In the third phase (Prediction), the narrative \(e_i\) directly drives the decision.  After generating \(e_i\), the model issues its prediction $\hat y_i$ by
\[
\hat y_i \sim \pi_\theta\bigl(\,\cdot\mid x_i, c_i, e_i\bigr),
\]
presented as ``Approve'' or ``Deny.'' Because the explanation precedes the decision, accuracy rewards flow back through the explanation tokens that set up correct outcomes.

\subsection{Reflection-augmented Supervised Fine-tuning}
Before reinforcement learning, we insert a supervised fine-tuning (SFT) stage to serve two purposes. First, it raises baseline predictive performance and calibration by warming the model to domain vocabulary, feature semantics, and label definitions. Second, it establishes the fixed Reasoning to Explanation to Prediction generation order and enforces schema-consistent outputs under both expert and consumer prompts. This strategy is a standard precursor to reinforcement learning in the literature \citep{ouyang_training_2022}.

For SFT, we require instruction-response pairs. Curating human instruction–response datasets for mortgage decisions would be ideal but is costly at scale. We therefore use instruction–response pairs produced by a strong reference model (GPT-5), which provides reliable targets for format and content tuning at a low cost.

For each input \(x_i\) and audience prompt, the reference model emits a structured response that contains a concise explanation \(e_i^{\mathrm{ref}}\) and a decision \(\hat y_i^{\mathrm{ref}}\). The reference outputs do not include internal reasoning traces, so there is no chain-of-thought in the targets. If the initial reference decision disagrees with the ground truth \(y_i\), we trigger a single reflection pass that asks the reference model to diagnose the likely mistake and regenerate the final response; this recovers harder cases and avoids biasing the SFT set toward only easy examples. Reflection augmentation is necessary because it increases target quality and coverage on difficult files, which improves calibration and reduces overfitting to superficial patterns.

We mask the response and train the Qwen3-4B model with token-level cross-entropy to reproduce \(e_i^{\mathrm{ref}}\) and \(\hat y_i^{\mathrm{ref}}\) under the three-phase generation schema. The chain \(c_i\) is latent during SFT and is not supervised. This stage yields a format-consistent initialization for both expert and consumer prompts and improves the baseline predictive performance, which enhances the efficiency and stability of the subsequent GRPO stages.

\subsection{Two Stages of GRPO Fine-tuning}
\subsubsection{An overview of Group Relative Policy Optimization.}
\begin{figure}[htb!]
			\centering
			\caption{Details on Group Relative Policy Optimization}
			\includegraphics[width=0.8\linewidth]{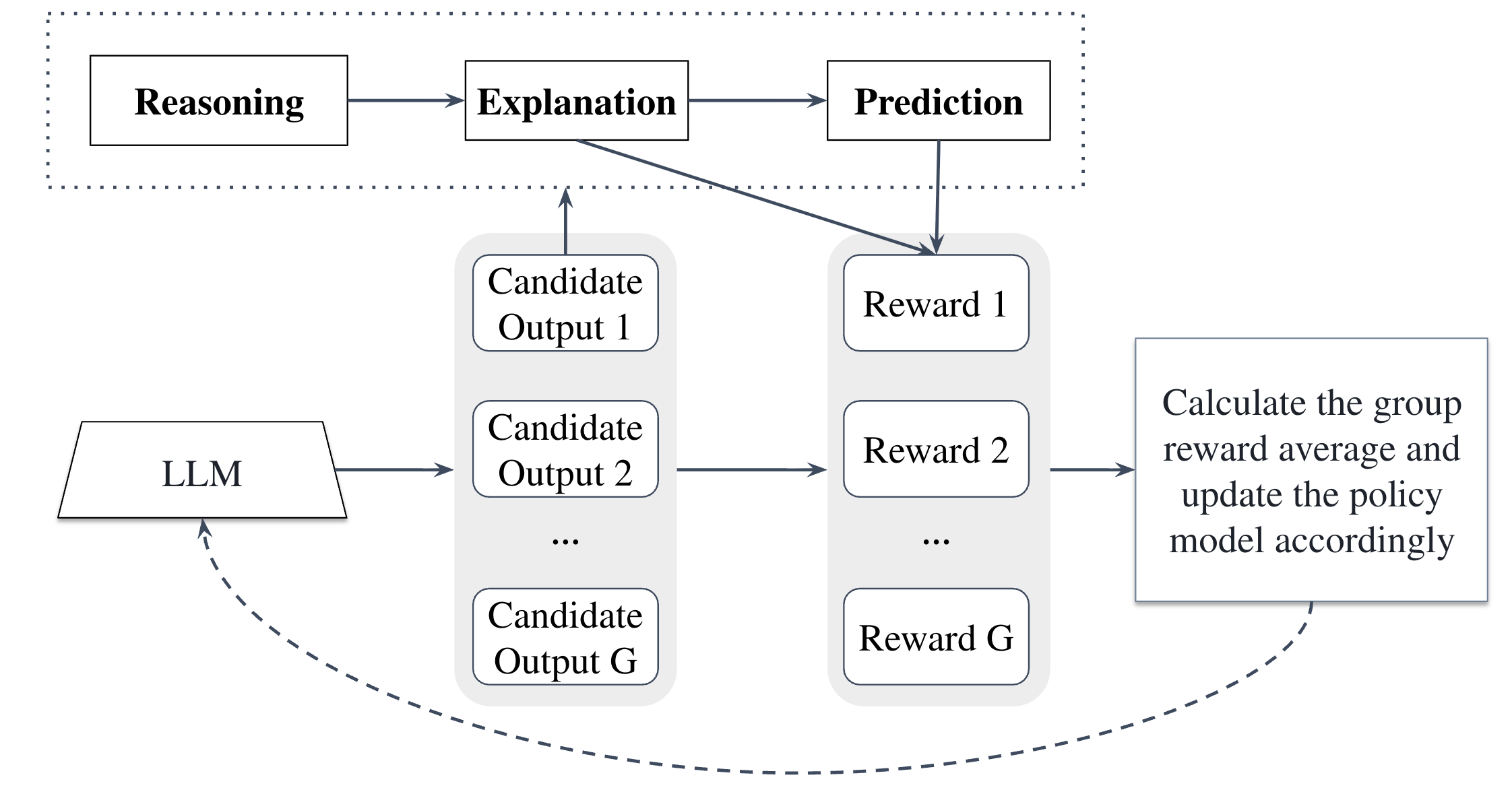}
			\begin{flushleft}
				\footnotesize \parbox{1.0\linewidth}{
					\textit{Notes.} Three‐phase generation pipeline (Reasoning, Explanation, Prediction) with GRPO fine-tuning to jointly optimize predictive accuracy and explanation quality.
				}
			\end{flushleft}
			\label{fig:grpo_detail}
		\end{figure}

To optimize behavior along the exact sequence used at inference, we apply GRPO. The algorithm samples multiple complete trajectories per input, computes group-relative advantages, and applies a clipped surrogate objective with KL regularization to stabilize updates \citep{guo_deepseek-r1_2025}. Figure \ref{fig:grpo_detail} demonstrates the fine-tuning logic of GRPO combined with the three-phase generation requirement.
We sample \(G\) complete trajectories per input using the previous policy \(\pi_{\theta_{\mathrm{old}}}\), compute group-relative advantages, and update the current policy with a clipped, KL-regularized surrogate. Here \(G\) is the number of trajectories in the group, \(\epsilon\) controls clipping of probability ratios, and \(\beta\) scales the KL term that keeps \(\pi_\theta\) close to \(\pi_{\theta_{\mathrm{old}}}\).

Specifically, for each example \((x_i,y_i)\) the procedure samples \(G\) full Reasoning–Explanation–Prediction trajectories, denoted as $o_{i,j}$, using the old policy \(\pi_{\theta_\mathrm{old}}\),
\[
o_{i,j} \,=\, \bigl(c_{i,j}, e_{i,j},\,\hat y_{i,j}\bigr) \quad (j=1,\dots,G).
\]
Each trajectory yields a reward $r_{i,j}$. This reward can reflect decision correctness or explanation tone–related dimensions; details are provided in the next two subsections.

The group baseline and relative advantage are computed as
\[
\bar r_i \,=\, \frac{1}{G}\sum_{j=1}^G r_{i,j},
\qquad
\widehat A_{i,j} \,=\, r_{i,j} - \bar r_i.
\]
Let \(\rho_{i,j} = \frac{\pi_\theta(o_{i,j})}{\pi_{\theta_{\mathrm{old}}}(o_{i,j})}\) be the trajectory-level probability ratio, where \(\pi_\theta(o_{i,j})\) is the product of token probabilities across Reasoning, Explanation, and Prediction. The GRPO surrogate is
\[
J_{\mathrm{GRPO}}(\theta)
=\mathbb{E}\!\Biggl[\frac{1}{G}\sum_{j=1}^G
\min\!\bigl(\rho_{i,j}\,\widehat A_{i,j},\;\mathrm{clip}(\rho_{i,j},1-\epsilon,1+\epsilon)\,\widehat A_{i,j}\bigr)\Biggr]
-\beta\,D_{\!KL}\bigl(\pi_\theta\;\|\;\pi_{\theta_{\mathrm{old}}}\bigr).
\]
Clipping, controlled by \(\epsilon\), limits the change in probability ratios, and KL regularization, scaled by \(\beta\), penalizes divergence from the reference policy. Maximizing \(J_{\mathrm{GRPO}}(\theta)\) increases the likelihood of high-reward trajectories and suppresses low-reward ones. Because the decision is conditioned on the explanation, accuracy rewards implicitly select for explanations that support correct decisions.

\subsubsection{GRPO tuning for decision correctness.}

GRPO-Step1 starts from the SFT checkpoint, attaches the ACC adapter, and freezes all base parameters. Training prompts alternate between an expert format and a consumer format. The alternation is intentional because mortgage decisions will be queried by different audiences using different phrasings. Alternating formats during training forces the model to produce decisions that are consistent across audiences, which reduces framing sensitivity.

For each \((x_i,y_i)\) the GRPO procedure samples \(G\) trajectories from \(\pi_{\theta_\mathrm{old}}\) under the current audience format, computes \(r_{i,j}=\mathbf{1}\{\hat y_{i,j}=y_i\}\), group baselines, and advantages, and then optimizes a GRPO objective over complete trajectories while updating only the ACC adapter. Gradients flow through the entire Reasoning–Explanation–Prediction trajectory, since correctness depends on how reasoning and explanation shape the final label. For inference, expert prompts use this GRPO-Step1 checkpoint.

\subsubsection{GRPO tuning for explanation tone.}

GRPO-Step2 attaches the TONE adapter while keeping the ACC adapter active for generation and frozen for updates. Only the TONE adapter is trained, and the reward is computed based on consumer explanations only.

We measure readability using the Flesch–Kincaid Grade Level, a rule-based formula that depends on average sentence length and average syllables per word. Let \(\mathrm{FK}(e_{i,j})\) denote the Flesch–Kincaid Grade Level of explanation \(e_{i,j}\), then the readability reward is \(r_{\text{read}}(e_{i,j})=\mathbf{1}[\text{FK}(e)\le 8]\), which targets plain language suitable for broad consumer audiences and yields a binary reward. We choose grade 8 because financial messages to consumers should be easily understood by borrowers with diverse literacy backgrounds; this level is widely used as a plain English benchmark that balances accessibility with the need to convey precise information. This dictionary- and rule-based approach is widely used in the literature (e.g., \citealp{loughran_measuring_2014}) and is fast to compute.

We measure politeness using the rule-based density of politeness markers computed with Stanford's politeness strategy taxonomy \cite{danescu-niculescu-mizil_computational_2013}. For each generated explanation $e$, we extract spans that match predefined politeness strategies such as gratitude expressions (e.g., ``thank you,'' ``appreciate'') and deference markers (e.g., ``please''), then compute a token-level density $d(e_{i,j})$.
We multiply this density by four and cap the result at 1 so that the politeness reward shares the same $[0,1]$ range as the readability reward and the two objectives have comparable weight.
This also reflects the goal that we encourage messages that are polite but not overloaded with formulaic cues. Specifically, the politeness reward is \(r_{\text{polite}}(e_{i,j})=\min(1,4\,d(e_{i,j}))\).

The tone reward in GRPO-Step2 is the sum of these two components, \(r_{i,j}=r_{\text{read}}(e_{i,j})+r_{\text{polite}}(e_{i,j})\). For each \((x_i,y_i)\) the GRPO procedure samples \(G\) trajectories from \(\pi_{\theta_\mathrm{old}}\) under the consumer prompt, computes reward, and then optimizes a GRPO objective over complete trajectories while updating only the TONE adapter and the ACC adapter remains frozen. This increases the likelihood of explanations that are readable and polite without changing the decision distribution.
For inference, consumer prompts use the GRPO-Step2 checkpoint.

\section{Evaluation Strategy} \label{sec:eval_strategy}

\subsection{Dataset}
We evaluate the LEXMA framework on a real-world dataset. Mortgage origination is a critical setting in which AI increasingly supports high-stakes decision making. We therefore use the Home Mortgage Disclosure Act (HMDA) public Loan Application Register, accessible through the CFPB HMDA data portal, a nationwide corpus of application-level mortgage records reported by covered U.S. financial institutions.\footnote{Data access: \url{https://ffiec.cfpb.gov/data-publication/}. Data field descriptions: \url{https://ffiec.cfpb.gov/documentation/publications/loan-level-datasets/lar-data-fields}.} The public HMDA files include standardized loan and applicant attributes such as loan amount, interest rate, loan purpose, lien status, occupancy type, property value, applicant income, debt-to-income ratio, combined loan-to-value ratio, and action taken. They also append census tract characteristics and demographic details.

To mitigate potential LLM memorization concerns, we restrict attention to the 2024 subset. We retain only applications with \texttt{action\_taken} equal to 1 (loan approved) or 3 (loan denied) and discard other statuses, including approved but not accepted and withdrawn. Based on the field definitions, we drop variables that are undefined before the credit decision or populated only when a loan is originated, such as total loan costs, total points and fees, and lender credits. We also exclude protected-class attributes and their derivatives (race, ethnicity, sex) to conform with fair-lending law. After these filters, the training dataset contains 41 input variables, and the outcome is a binary label indicating approval versus denial. Approximately 75\% of applications are approved. The processed dataset includes more than one million rows, although we fine-tune our model using only a small fraction of the available observations.
To mitigate the impact of this class imbalance during model training, we construct class-balanced subsets for all fine-tuning stages. Specifically, for the supervised fine-tuning and both GRPO stages, we oversample denied applications so that the training data used in each stage contains 50\% approved and 50\% denied cases.

For preprocessing, since our inputs are tabular, we need to apply certain techniques to make them more readable for LLMs. Studies have shown transforming tables into LLM-readable text helps the model exploit column semantics \citep{fang_large_2024} and improve predictive performance \citep{hegselmann_tabllm_2023}. Therefore, we serialize each HMDA record into a compact narrative using a template-based approach.

\subsection{Fine-tuning Experimental Setup}

\subsubsection{Prompts for the generation task.}
We adopt the following zero-shot prompt templates for the generation task and apply them in all stages of LEXMA fine-tuning. The only exception is the reflection-augmented component, for which the prompts are provided in the Appendix.

{ \OneAndAHalfSpacedXI
\begin{tcolorbox}[
    colframe=black,
    colback=gray!10,
    fonttitle=\bfseries,
    sharp corners=all,
    title=Prompt template for loan approval decision explanations shown to experts,
    width=\textwidth,
    boxrule=1pt,
    left=5pt,
    right=5pt,
    top=5pt,
    bottom=5pt
]
\textbf{System Role:}

You are a skilled loan expert. Your task is to evaluate the following loan application and determine whether to approve or deny the loan. Provide a concise, expert-facing justification for the decision. Always respond strictly in the requested JSON format.

\textbf{User Role:}

Here is the loan information: \{summary\}

(END OF LOAN INFORMATION)

Respond only with JSON in this exact format: \texttt{\{"Officer\_Message": "<A short paragraph for the loan officer explaining and justifying the decision>", "Decision": "<Approved or Denied>"\}}.

Now start!
\end{tcolorbox}
}

{ \OneAndAHalfSpacedXI
\begin{tcolorbox}[
    colframe=black,
    colback=gray!10,
    fonttitle=\bfseries,
    sharp corners=all,
    title=Prompt template for loan approval decision explanations shown to consumers,
    width=\textwidth,
    boxrule=1pt,
    left=5pt,
    right=5pt,
    top=5pt,
    bottom=5pt
]
\textbf{System Role:}

You are a skilled loan expert communicating directly with the applicant. Write in a polite, encouraging tone with clear, jargon-free language. Keep explanations actionable and easy to understand. If approving: summarize what went well and offer 1–2 simple tips for maintaining strong credit. If denying: list main reasons in plain language and give some specific, doable steps the applicant can take to improve their chances next time.

\textbf{User Role:}

Here is the loan information: \{summary\}

(END OF LOAN INFORMATION)

Write a short, friendly note directly to the applicant in plain language.

Respond only with JSON in this exact format: \texttt{\{"Applicant\_Message": "<Your short note to the applicant with the guidance above>", "Decision": "<Approved or Denied>"\}}.

Now start!
\end{tcolorbox}
}

\subsubsection{Implementation details.}
We select the Qwen3-4B model for its strong balance between performance and computational efficiency, making it well-suited for reinforcement learning with limited resources. Qwen3-4B is an open-source foundation model that has demonstrated competitive reasoning and generation quality across a variety of benchmarks \citep{qwen_team_qwen3_2025}.
To establish a baseline for comparison, we prompt the Qwen3-4B model to perform the loan approval prediction and explanation task using the same prompt employed during the fine-tuning process.

To test hyper-parameter settings in GRPO, we varied the group size \(G\) (the number of trajectories sampled per case) between 8 and 16 and explored learning rates of \(5\times 10^{-4}\), \(2\times 10^{-4}\), and \(1\times 10^{-4}\). We find that increasing \(G\) from 8 to 16 produced only marginal improvements in predictive metrics while approximately doubling training time. A learning rate of \(5\times 10^{-4}\) was too aggressive and reduced performance, whereas \(2\times 10^{-4}\) and \(1\times 10^{-4}\) performed similarly; \(2\times 10^{-4}\) reached its peak in fewer training steps. We therefore adopt a learning rate of \(2\times 10^{-4}\) for the main experiments. The maximum generation length is set to 1{,}024 tokens. We set temperature to be 1 during GRPO training.

All fine-tuning experiments run on two NVIDIA RTX A6000 GPUs with the Qwen3-4B model loaded in 8-bit precision and trained in mixed-precision FP16. All components are trained via Parameter-Efficient Fine-Tuning (PEFT) using low-rank adapters (LoRA), freezing all other weights to minimize memory and compute overhead \citep{hu_lora_2022}.\footnote{For the ACC adapter we modify the attention projections (query, key, value, output) and the gating projection. For the TONE adapter we modify the MLP up- and down-projection layers.}
 We accumulate gradients over 64 steps with a per-device batch size of one.
We report the training data size and training time: SFT uses 30{,}000 loan cases and takes about 6 hours; GRPO-Step1 uses 15{,}000 cases and takes about 30 hours; and GRPO-Step2 uses 1{,}000 cases and takes about 2 hours. The SFT training data and the GRPO training data are disjoint, and the held-out test data is not used for SFT or GRPO.

\subsection{Performance Evaluation Protocols}
This subsection explains how we assess the system along two dimensions: decision correctness and explanation quality. We first describe how predictive performance is measured, then outline how explanation quality is evaluated for expert and consumer audiences.

\subsubsection{Decision correctness evaluation.}

To evaluate the predictive performance of LEXMA, we benchmark it against several representative alternatives: four machine learning models including XGBoost, Neural Network, Logistic Regression, and Gradient Boosting, and two LLM baselines, the raw Qwen3-4B model, and GPT-5 at a median reasoning intensity. All systems are evaluated on the same held-out test dataset under both expert and consumer prompts.\footnote{For the LLM-based prediction, the expert and consumer prompts induce distinct decision and explanation outputs. In contrast, XGBoost and the Neural Network produce a single prediction per case that does not depend on the audience, so their performance metrics are identical in the expert and consumer panels and are repeated only for ease of comparison.} Expert results correspond to the GRPO-Step1 checkpoint, while consumer results correspond to GRPO-Step2, since GRPO-Step2 tunes tone with the correctness adapter kept fixed.
Specifically, for XGBoost, we tune the number of trees, maximum tree depth, learning rate, and L2 regularization strength using grid search. For Neural Network classifier, we use a feedforward multilayer perceptron and tune the hidden-layer configuration, learning rate, and L2 regularization. For Logistic Regression, we tune the inverse regularization strength and maximum number of iterations. For Gradient Boosting baseline, we tune the number of trees, learning rate, maximum depth of individual trees, and subsampling rate.
For the two LLM baselines, raw Qwen3-4B and GPT-5, we set the temperature to be 1. For our method we set temperature to be 0. The lower temperature provides deterministic outputs, which removes sampling variance and yields stable results. At the time of the experiments, the GPT-5 API only supported temperature of 1, so we set the raw Qwen baseline to the same value for parity.

To isolate the sources of improvements, we conduct an ablation that evaluates four checkpoints from the training pipeline (raw Qwen, SFT, GRPO-Step1, GRPO-Step2), each assessed under expert and consumer prompts on the same held-out test cases.

\subsubsection{Explanation quality evaluation.}

To evaluate explanation quality, we mainly rely on IRB-approved human evaluation. Assessment from human evaluators is important because the system is built to serve two real audiences and only those audiences can judge whether the explanations are truly useful. Loan professionals assess expert-facing notes and prospective applicants assess consumer-facing explanations, allowing us to measure perceived clarity, relevance, fairness, actionability, and trust directly from end users. In addition, we report a mix of quantitative and qualitative evidence, which together provide a holistic view of explanation quality.

\paragraph{Human evaluation for expert explanations.}
Expert-facing quality is assessed with a small but highly experienced panel. We recruit three loan professionals who are currently working in the mortgage origination industry, two with more than twenty years of experience and one with fourteen years. This panel reflects the audience that would consume expert-facing rationales in practice.

Each expert reviews ten mortgage applications, yielding thirty paired evaluations. For each case, the expert reads the file summary and two anonymized explanations generated for the same input, one from the raw Qwen3-4B model and one from the fine-tuned model, presented in randomized order. We first record an overall choice between the two explanations for each case. We then ask the expert to rate each message on three 1–5 Likert scales: risk relevance (whether the rationale focuses on truly decision-relevant factors), decision appropriateness (fit between the rationale and the proposed decision for the file), and explainability (how well the message conveys why the decision was reached). After completing the cases, experts answer brief questions on comparative utility and mental effort relative to their current practices.

\paragraph{Human evaluation for consumer explanations.}
We recruit 110 workers on Prolific and ask each participant to rate five mortgage application pairs. Each mortgage application is rated by at least four raters. Participants read a short application summary and two versions of the consumer-facing decision message for the same case,  one from the raw Qwen3-4B model and one from the fine-tuned model, presented in randomized order. Then the participants are asked to choose which message they would prefer to receive if they were the consumer, and rate each message on seven dimensions: clarity (is the message clear and easy to understand), politeness (is the tone respectful and courteous), fairness (does the decision seem fair given the information), actionability (are the next steps concrete and doable), reasonableness (are the message and its reasons sensible), trustworthiness (are the message and its reasons credible), and satisfaction (overall satisfaction with the message), each on 1–5 Likert scales. We exclude two workers' results for short completion times, and obtain 536 rater-loan pairs.

For consumer explanations, we also quantify the two reward dimensions directly tuned in our workflow. Specifically, we examine the distribution of readability grades and politeness densities to see whether consumer-facing explanations move toward plain language and whether tone rises to a polite level.

\section{Evaluation Results}\label{sec:eval_result}

\subsection{Improvement in Decision Correctness}
\begin{figure}[htb!]
			\centering
			\caption{Decision Correctness Comparison}
			\includegraphics[width=1\linewidth]{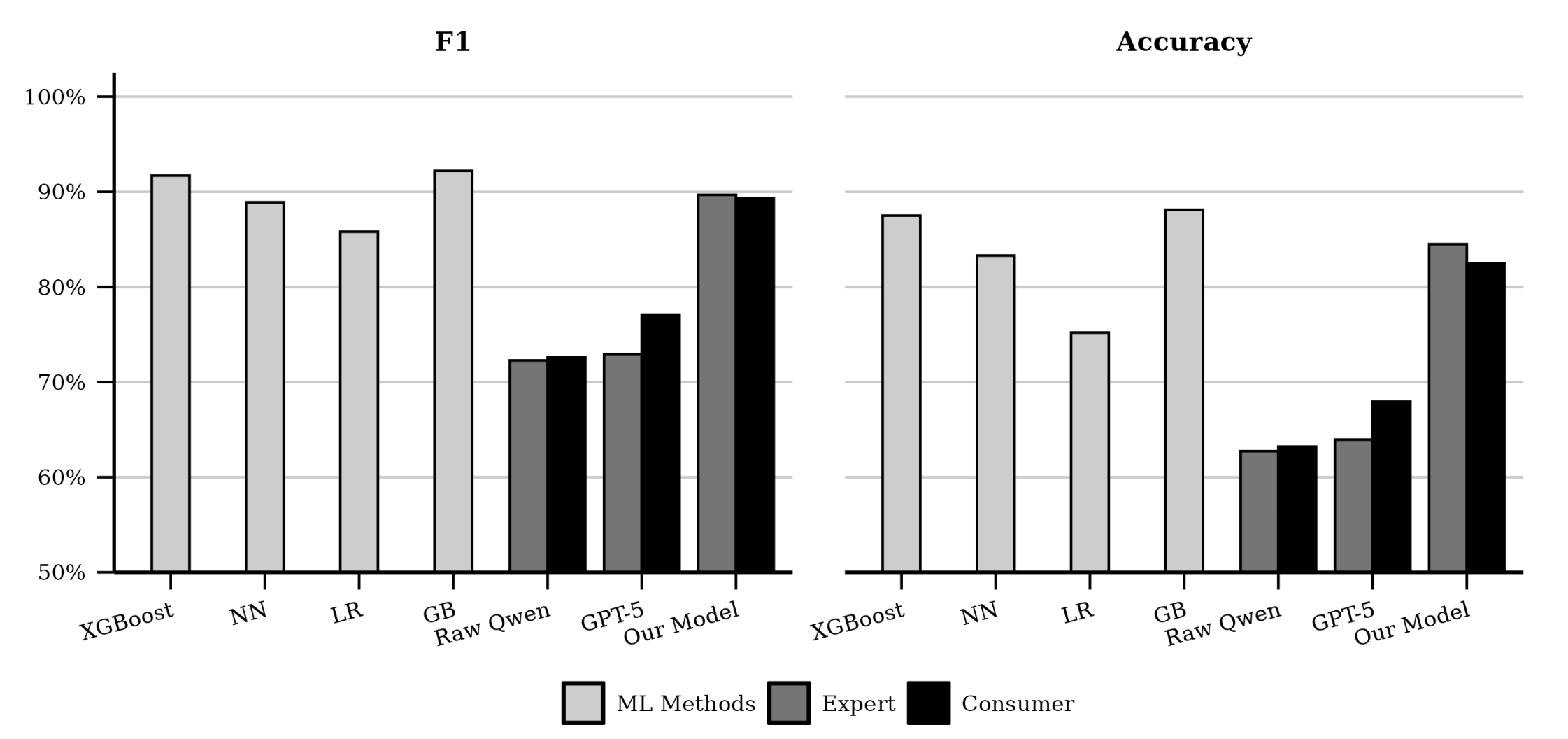}
			\begin{flushleft}
				\footnotesize \parbox{1.0\linewidth}{
					\textit{Notes.} NN: Neural Network. LR: Logistic Regression. GB: Gradient Boosting. In our method, expert results are generated on the test set using the GRPO-Step1 checkpoint, while consumer results use GRPO-Step2.
    }
			\end{flushleft}
			\label{fig:performance_benchmark}
		\end{figure}

We benchmark LEXMA on the HMDA test set against four machine learning baselines (XGBoost, Neural Network, Logistic Regression, and Gradient Boosting), the raw Qwen3-4B model, and GPT-5 at a median reasoning intensity. Figure \ref{fig:performance_benchmark} and Table~\ref{tab:predictive_performance} summarize the results.
Compared with the two LLM baselines, LEXMA shows substantial gains in decision correctness. Under the expert prompt, our method reaches F1 of 0.897 and accuracy of 0.845, improving over raw Qwen at 0.723 and 0.627 and over GPT-5 at 0.730 and 0.639. Under the consumer prompt, our method achieves F1 of 0.893 and accuracy of 0.825, exceeding raw Qwen at 0.726 and 0.632 and GPT-5 at 0.771 and 0.679. These improvements are notable because GPT-5 is a substantially larger model than the Qwen3-4B backbone used in our fine-tuning. The result highlights the effectiveness of the proposed fine-tuning framework in ensuring decision correctness.

Our method attains predictive performance that is close to strong tree baselines such as XGBoost and Gradient Boosting, competitive with Neural Network, and better than Logistic Regression, while providing capabilities that these models do not offer.  This observation is consistent with prior evidence on tabular prediction, which shows that on prediction tasks based on tabular datasets, the gradient-boosted methods such as XGBoost often exceed LLM-based approaches \citep{grinsztajn_why_2022}. The contribution of our method is improving the predictive performance of LLMs, which achieves performance close to XGBoost, while also generating concise, audience-appropriate narrative explanations that are tightly coupled to each decision, whereas XGBoost relies on feature importance and template reason codes that do not provide comparable, case-specific narratives.

{\OneAndAHalfSpacedXI
\begin{table}[h!]
\small

\centering

\caption{Decision Correctness Comparison}

\label{tab:predictive_performance}

\begin{threeparttable}

\begin{tabular}{@{\extracolsep{\fill}}>{\raggedright\arraybackslash}p{0.10\textwidth}>{\centering\arraybackslash}p{0.08\textwidth}>{\centering\arraybackslash}p{0.08\textwidth}>{\centering\arraybackslash}p{0.08\textwidth}>{\centering\arraybackslash}p{0.08\textwidth}>{\centering\arraybackslash}p{0.08\textwidth}>{\centering\arraybackslash}p{0.08\textwidth}>{\centering\arraybackslash}p{0.08\textwidth}>{\centering\arraybackslash}p{0.08\textwidth}>{\centering\arraybackslash}p{0.08\textwidth}}

\toprule

    & \multicolumn{6}{c}{\textbf{Benchmarks}} & \multicolumn{3}{c}{\textbf{Our Method}} \\

    \cmidrule(lr){2-7} \cmidrule(lr){8-10}

    &  \textbf{XGBoost} & \textbf{NN}& \textbf{LR}& \textbf{GB}& \textbf{Raw Qwen} & \textbf{GPT-5} & \textbf{SFT} & \textbf{GRPO-Step1}& \textbf{GRPO-Step2}\\

\midrule

 \multicolumn{10}{l}{\textbf{Expert Prompt}} \\
 F1 & 0.917 & 0.889 &  0.858&   0.922& 0.723 & 0.730 & 0.825 & 0.897 & 0.902 \\

 Accuracy & 0.875 & 0.833 &  0.752&  0.881&  0.627 & 0.639 & 0.757 & 0.845 & 0.851 \\

 Precision & 0.922 & 0.895 & 0.756&  0.910&   0.828 & 0.844 & 0.908 & 0.903 & 0.900 \\

 Recall & 0.911 &  0.882&  0.993&  0.935&  0.641 & 0.642 & 0.756 & 0.890 & 0.903 \\

\midrule

 \multicolumn{10}{l}{\textbf{Consumer Prompt}} \\

 F1 & 0.917 & 0.889&  0.858&  0.922&  0.726 & 0.771 & 0.793 & 0.893 & 0.893 \\

 Accuracy & 0.875 &0.833  &  0.752&  0.881& 0.632 & 0.679 & 0.717 & 0.840 & 0.825 \\

 Precision & 0.922 &0.895 &  0.756&   0.910& 0.831 & 0.840 & 0.886 & 0.898 & 0.830 \\

 Recall & 0.911 &0.882  &  0.993&  0.935& 0.645 & 0.713 & 0.718 & 0.889 & 0.967 \\

\bottomrule

\end{tabular}

\begin{tablenotes}

    \footnotesize

    \setlength{\baselineskip}{\normalbaselineskip}

    \item \textit{Notes.} NN: Neural Network. LR: Logistic Regression. GB: Gradient Boosting.
    Expert-facing deployment uses the GRPO-Step1 checkpoint. The expert results reported for GRPO-Step2 are included only to verify that tone tuning preserves decision correctness. Consumer-facing deployment uses the GRPO-Step2 checkpoint. For XGBoost, NN, LR, and GB, there is no difference between expert and consumer prompts since these models produce a single prediction per case.

\end{tablenotes}

\end{threeparttable}

\end{table}
}

\begin{figure}[htb!]
			\centering
			\caption{Ablation Study}
			\includegraphics[width=1\linewidth]{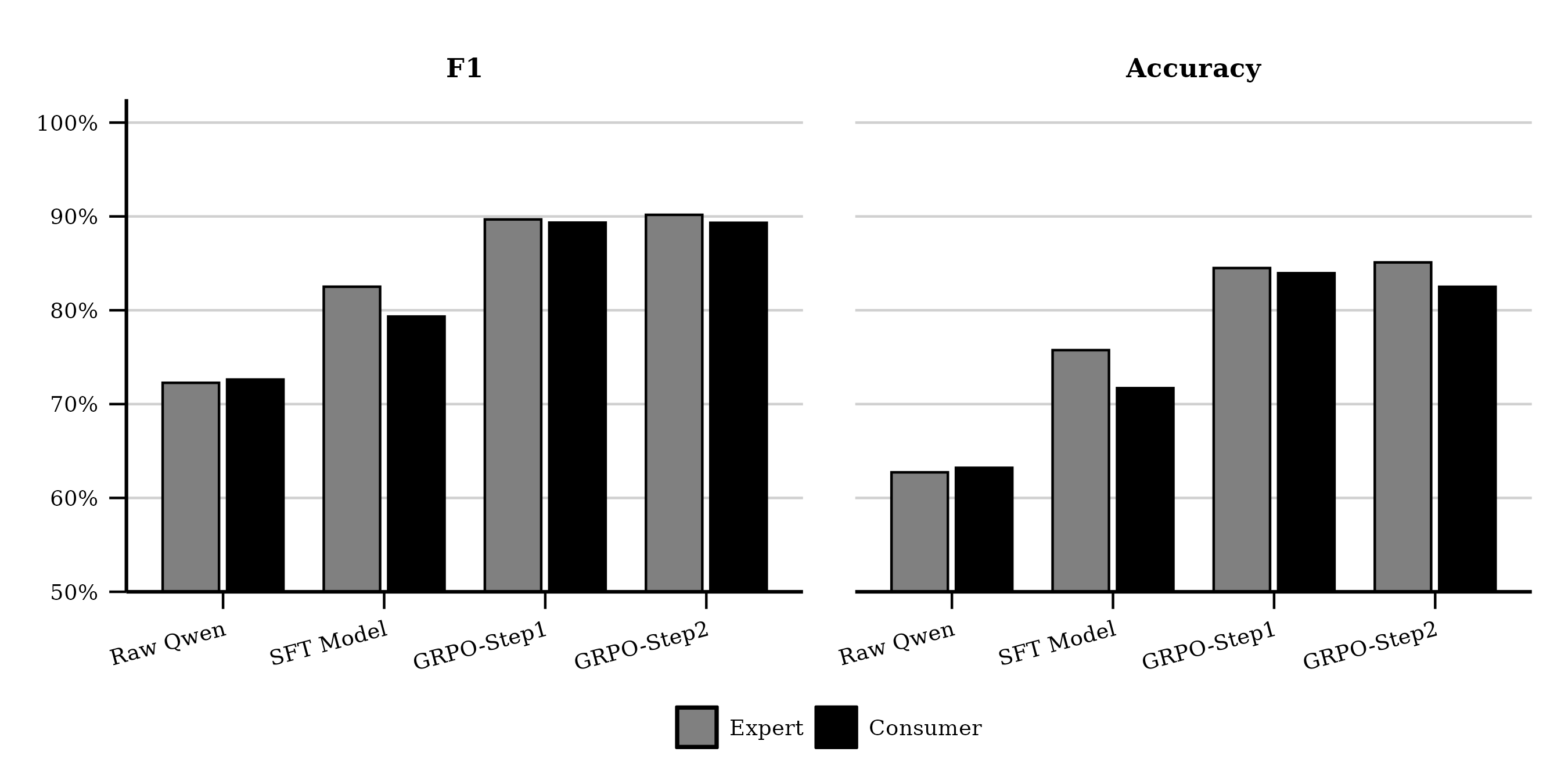}
			\begin{flushleft}
				\footnotesize \parbox{1.0\linewidth}{
					\textit{Notes.} Comparison of predictive performance across the proposed framework: reflection augmented supervised fine-tuning (SFT), GRPO-Step1 for decision correctness, and GRPO-Step2 for explanation tone tuning.
				}
			\end{flushleft}
			\label{fig:performance_ablation}
		\end{figure}

The ablation isolates which stages of the workflow contribute to the improvement in predictive performance. Figure~\ref{fig:performance_ablation} plots F1 and accuracy for four checkpoints that represent the training procedure in Section~\ref{sec:method}: the raw Qwen3-4B model, reflection-augmented supervised fine-tuning (SFT), GRPO-Step1 for decision correctness tuning, and GRPO-Step2 for explanation tone tuning. All checkpoints are evaluated on the same test dataset under both expert and consumer prompts, and the corresponding values appear in Table~\ref{tab:predictive_performance}.
Starting from the raw model, SFT yields clear improvements in decision correctness. Under the expert prompt, F1 rises from 0.723 to 0.825 and accuracy from 0.627 to 0.757. Under the consumer prompt, F1 increases from 0.726 to 0.793 and accuracy from 0.632 to 0.717. These gains reflect the benefit of structured targets and a single reflection pass before reinforcement learning.
GRPO-Step1 further optimizes decision correctness. Relative to SFT, expert F1 increases from 0.825 to 0.897 and accuracy from 0.757 to 0.845. On the consumer side, F1 increases from 0.793 to 0.893 and accuracy from 0.717 to 0.840.
GRPO-Step2 applies explanation tone tuning with the correctness adapter frozen, and the predictive metrics remain essentially unchanged, which is the intended outcome. For expert prompts, GRPO-Step2 is very close to GRPO-Step1 at F1 0.902 and accuracy 0.851.\footnote{We include the expert-side GRPO-Step2 results to demonstrate that explanation tone tuning does not degrade expert decision quality. Throughout, the GRPO-Step1 checkpoint remains the final expert model, while GRPO-Step2 is the final consumer model.} For consumer prompts, F1 remains 0.893 and accuracy is 0.825, a small decrease of 0.015 from GRPO-Step1. These results show that tone optimization reshapes style without materially altering the decisions, consistent with the separation of roles between the ACC and TONE adapters in our training design.

\subsection{Expert Explanation Quality Enhancement}
We first report the results from human evaluation. Across the 30 paired reviews, experts preferred the fine-tuned message in 24 cases, which corresponds to an 80\%  preference rate.
Expert ratings also favor the fine-tuned explanations on all reported dimensions.  As shown in Table~\ref{tab:human_eval_expert}, risk relevance rises from 3.4 to 4.4, decision appropriateness from 2.2 to 3.9, and explainability from 2.4 to 3.6, with all differences significant at the 5 percent level. On a five-point scale, these shifts are large: decision appropriateness increases by 1.7 points, moving from below the neutral midpoint of 3 to near agreement that the proposed decision fits the file; perceived explainability increases by 1.2 points, crossing the neutral threshold and indicating clearer reasons for the decision; and risk relevance improves by 1.0 point, from slightly above neutral to strong agreement that explanations focus on decision-critical factors.
Furthermore, experts noted that summaries were easy to read and could speed up review. Regarding the practical utility of having such an AI system and the cognitive load in processing files with AI explanations, two of the three experts reported that the AI explanations were somewhat more useful and slightly easier to work with than their current practices, while one reported no material difference.
{\OneAndAHalfSpacedXI
\begin{table}[h!]
\small
\centering
\caption{Expert Ratings on Raw versus Fine-tuned Explanations}
\label{tab:human_eval_expert}
\begin{threeparttable}
\begin{tabular}{@{\extracolsep{\fill}}>{\raggedright\arraybackslash}p{0.25\textwidth}>{\centering\arraybackslash}p{0.20\textwidth}>{\centering\arraybackslash}p{0.20\textwidth}>{\centering\arraybackslash}p{0.12\textwidth}>{\centering\arraybackslash}p{0.12\textwidth}}
\toprule
\textbf{Question} & \textbf{Raw Model Mean Score}& \textbf{Fine-tuned Model Mean Score}& \textbf{$t$}& \textbf{$p$}\\
\midrule
Risk Relevance& 3.400 & 4.400 & -2.466 & 0.026*  \\
Decision Appropriateness& 2.200 & 3.900 & -3.258 & 0.005** \\
Explainability& 2.400 & 3.600 & -3.182 & 0.006** \\
\bottomrule
\end{tabular}
\begin{tablenotes}
\footnotesize
\setlength{\baselineskip}{\normalbaselineskip}
\item Notes. Number of observations: 30. Ratings use 1 to 5 Likert scales where higher scores indicate better performance. Entries show item means by model, followed by two-sided \(t\)-tests.
 * $p < 0.05$, ** $p < 0.01$, *** $p < 0.001$.
\end{tablenotes}
\end{threeparttable}
\end{table}
}

\begin{figure}[htb!]
			\centering
			\caption{Demonstration of Enhanced Expert Explanation and Prediction Capability after Fine-tuning}
			\includegraphics[width=1\linewidth]{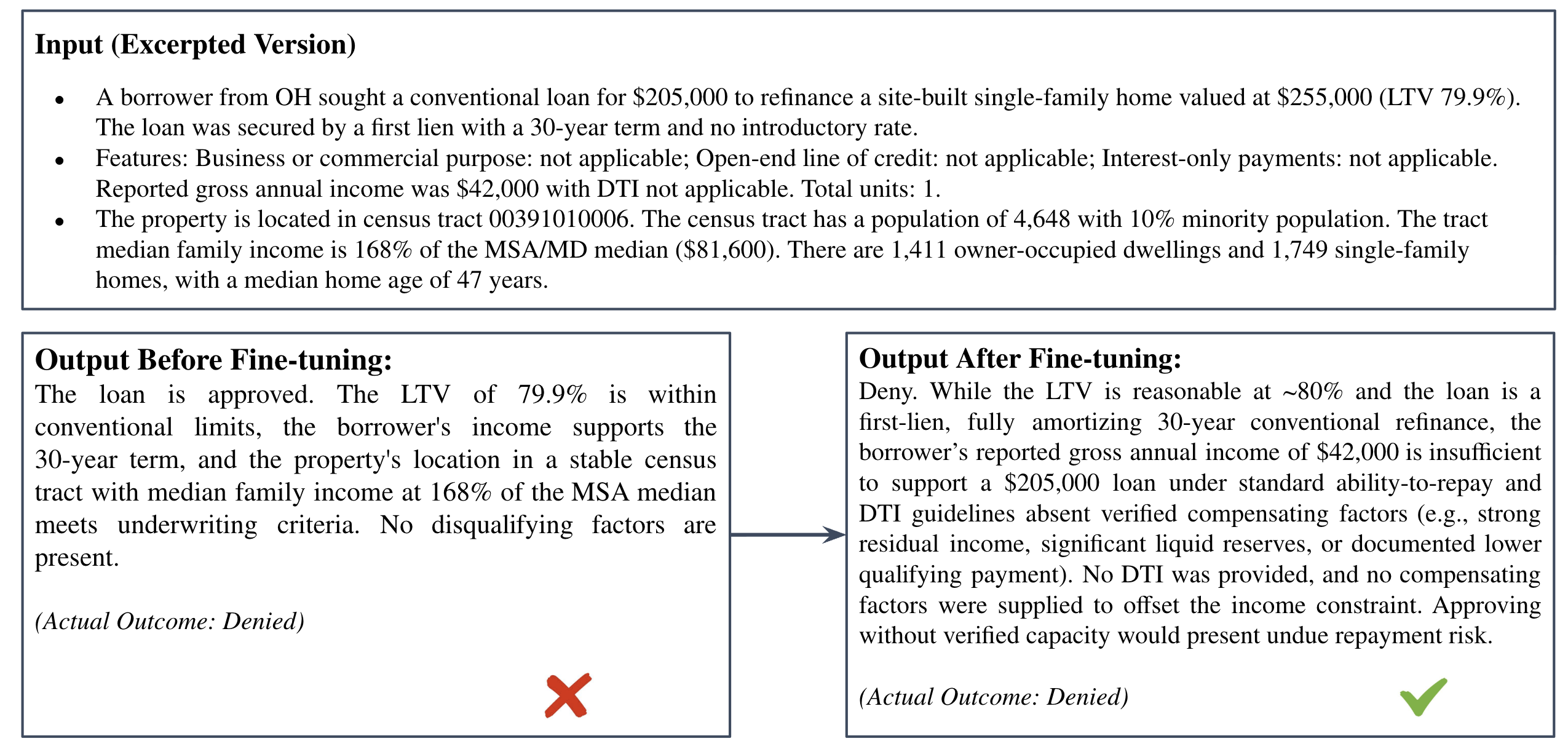}
			\begin{flushleft}
				\footnotesize \parbox{1.0\linewidth}{
					\textit{Notes.} A comparison between raw and fine-tuned models illustrates how fine-tuning improves explanatory quality by aligning the model's attention with relevant risk-indicative features.
				}
			\end{flushleft}
			\label{fig:compare_example_expert}
		\end{figure}

We further explore where the expert-explanation gains come from.  We hypothesize that this improvement stems from decision-correctness tuning that pushes the model to ground its rationales on case-specific evidence that actually produces the correct label. We conduct a qualitative review of predictions and accompanying narratives and find that the raw foundation model often produces plausible but superficial reasoning. Such behavior can be viewed as a form of hallucination: the model generates coherent and logical-sounding text that nonetheless fails to substantiate the prediction.
In contrast, the fine-tuned model exhibits a markedly improved ability to anchor its explanations in case-specific evidence. It highlights features that are temporally and contextually salient, forming explanations that are not only coherent but also aligned with the ground truth. This shift results in higher-quality reasoning and improved predictive accuracy.
A representative case in Figure~\ref{fig:compare_example_expert} illustrates this shift. This loan is denied in the real world. The raw model approves the loan while overlooking missing and insufficient capacity evidence; the fine-tuned model denies and explains that income at this level cannot support the obligation under ability-to-repay and debt-to-income guidelines absent verified offsets, which matches the actual outcome (Denied) and illustrates the shift from generic heuristics to concise, causally aligned justifications.

\subsection{Consumer Explanation Quality Enhancement}
\begin{figure}[htb!]
			\centering
			\caption{Improvement in Readability and Politeness in Consumer Explanations}
			\includegraphics[width=0.85\linewidth]{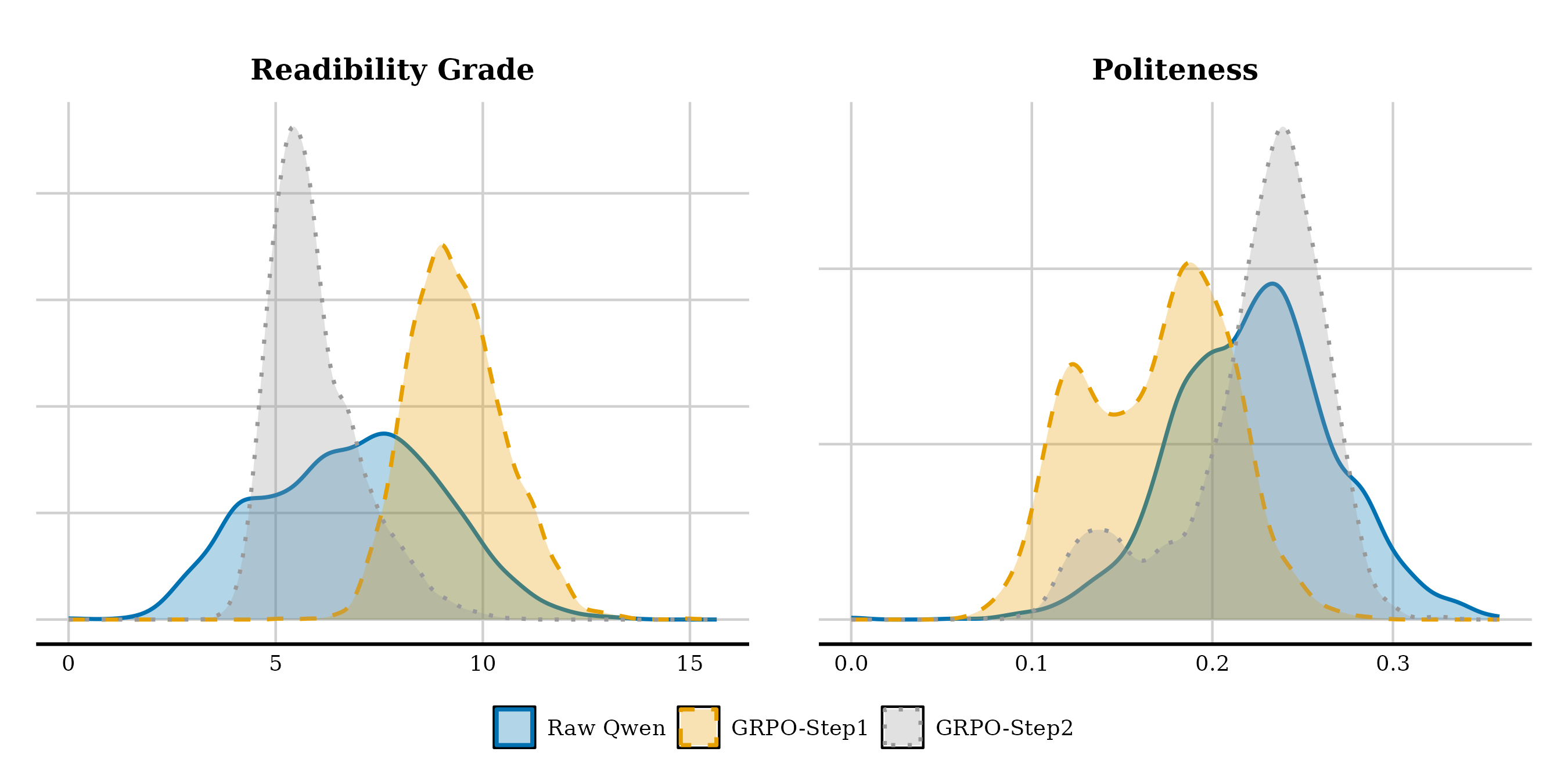}
			\begin{flushleft}
				\footnotesize \parbox{1.0\linewidth}{
					\textit{Notes.} Kernel density plots showing the distribution of readability grade (left panel) and politeness density (right panel) for three models: raw Qwen (baseline), GRPO-Step1 (after decision correctness tuning), and GRPO-Step2 (after explanation tone tuning). Each curve represents the probability density function for that model's outputs on the test dataset.   Lower readability grade indicates higher readability. Higher politeness density indicates higher politeness.
				}
			\end{flushleft}
			\label{fig:distribution_density_read_polite}
		\end{figure}

We first examine the two style dimensions directly tuned in our workflow. Figure~\ref{fig:distribution_density_read_polite} plots the distributions of readability grade and politeness density for raw Qwen, GRPO-Step1, and GRPO-Step2. Table~\ref{tab:statistics_comparison_read_polite} reports summary statistics for the same outputs. Readability is better when the grade is lower, with a target at or below grade eight for consumer explanations. Politeness density reflects the proportion of tokens covered by politeness markers, with reward saturation at 0.25 due to reward scaling.

We first observe that there is a clear shift after decision correctness tuning (GRPO-Step1). Relative to raw Qwen, GRPO-Step1 moves the readability distribution to the right and the politeness distribution to the left on consumer explanations. Table~\ref{tab:statistics_comparison_read_polite} shows the mean readability grade rising from 6.866 to 9.374, and the mean politeness density dropping from 0.223 to 0.169. Figure~\ref{fig:distribution_density_read_polite} reflects this pattern through a mass of Step1 samples around grades nine to ten and a thinner tail in the higher politeness range. This is potentially a side effect of optimizing for prediction correctness, and it motivates the subsequent explanation tone tuning.

Explanation tone tuning in GRPO-Step2 reverses these movements in the intended direction. Readability shifts left toward plain language, with the mean grade at 5.952 and the median at 5.718, which places most consumer explanations at or below the grade eight target. Politeness shifts right relative to Step1 and returns to a mean of 0.223. This level sits just below the 0.25 saturation point implied by the scaling, which encourages polite but non-formulaic phrasing. The standard deviations in Table~\ref{tab:statistics_comparison_read_polite} indicate that Step2 also concentrates outputs within a narrow band around the desired region for both measures.
These distributional results show that decision correctness tuning alone can unintentionally degrade consumer readability and politeness, while the subsequent explanation tone tuning restores both dimensions toward their targets.

{\OneAndAHalfSpacedXI
\begin{table}[h!]
\small

\centering

\caption{Readability and Politeness Comparison for Consumer Explanations}

\label{tab:statistics_comparison_read_polite}

\begin{threeparttable}

\begin{tabular}{@{\extracolsep{\fill}}>{\raggedright\arraybackslash}p{0.15\textwidth}>{\centering\arraybackslash}p{0.11\textwidth}>{\centering\arraybackslash}p{0.11\textwidth}>{\centering\arraybackslash}p{0.11\textwidth}>{\centering\arraybackslash}p{0.11\textwidth}>{\centering\arraybackslash}p{0.11\textwidth}>{\centering\arraybackslash}p{0.13\textwidth}}

\toprule

    & \textbf{Mean} & \textbf{Std. Dev.} & \textbf{Min} & \textbf{Max} & \textbf{Median} & \textbf{Mean Diff} \\

\midrule

 \multicolumn{7}{l}{\textbf{Readability Grade}} \\

 Raw Qwen & 6.866 & 2.136 & 0.000 & 13.360 & 6.981 & -- \\

 GRPO-Step1 & 9.374 & 1.154 & 5.148 & 15.647 & 9.272 & 2.509*** \\

 GRPO-Step2 & 5.952 & 1.083 & 3.643 & 10.815 & 5.718 & -3.423*** \\

\midrule

 \multicolumn{7}{l}{\textbf{Politeness Density}} \\

 Raw Qwen & 0.223 & 0.044 & 0.000 & 0.359 & 0.224 & -- \\

 GRPO-Step1 & 0.169 & 0.040 & 0.068 & 0.287 & 0.175 & -0.055*** \\

 GRPO-Step2 & 0.223 & 0.041 & 0.070 & 0.329 & 0.233 & 0.054*** \\

\bottomrule

\end{tabular}

\begin{tablenotes}

    \footnotesize

    \setlength{\baselineskip}{\normalbaselineskip}

    \item \textit{Notes.}  Lower readability grade indicates higher readability. Higher politeness density indicates higher politeness.  Mean difference is relative to the previous model in sequence.  * $p<0.10$, ** $p<0.05$, *** $p<0.01$.

\end{tablenotes}

\end{threeparttable}

\end{table}
}

We further evaluated consumer-facing explanations with a sample that mirrors the audience likely to receive such communications. Participants show a clear directional preference for the fine-tuned explanations, selecting them in 418 of 536 comparisons (78\%).
Fine-tuned explanations are perceived to be better than raw explanations in all seven dimensions. Clarity rises from 3.835 to 4.212 ($+0.377$) and trustworthiness from 3.742 to 4.139 ($+0.397$), indicating that consumers not only understand the fine-tuned message more easily but also view it as more credible. Actionability shows one of the largest movements, from 3.444 to 4.052 ($+0.608$), which suggests that the guidance becomes concretely actionable rather than generic. Satisfaction increases from 3.692 to 4.447 ($+0.755$), capturing the combined effect of clearer, kinder, and more useful explanations. Politeness rises from 4.124 to 4.586 ($+0.462$) and fairness from 4.180 to 4.551 ($+0.371$), pointing to explanations that feel both respectful and substantively justified. Reasonableness improves from 3.764 to 4.216 ($+0.452$), aligning perceptions of logic and evidence with the decision presented. On a five point scale, these absolute gains represent approximately 9 to 12 percent of the full scale width for clarity, fairness, reasonableness, and trustworthiness, about 15 percent for actionability, and nearly 19 percent for satisfaction. All fine-tuned means exceed 4.0, which places typical judgments in the agree range rather than hovering near neutrality, and all differences are statistically significant at $p<0.001$.

{\OneAndAHalfSpacedXI
\begin{table}[h!]
\small
\centering
\caption{Consumer Ratings on Raw versus Fine-tuned Explanations}
\label{tab:human_eval_consumer}
\begin{threeparttable}
\begin{tabular}{@{\extracolsep{\fill}}>{\raggedright\arraybackslash}p{0.20\textwidth}>{\centering\arraybackslash}p{0.14\textwidth}>{\centering\arraybackslash}p{0.14\textwidth}>{\centering\arraybackslash}p{0.16\textwidth}>{\centering\arraybackslash}p{0.16\textwidth}}
\toprule
\textbf{Question} & \textbf{Raw Model Mean Score}& \textbf{Fine-tuned Model Mean Score}& \textbf{$t$}& \textbf{$p$}\\
\midrule
Clarity& 3.835 & 4.212 & -6.640  & 0.000*** \\
Politeness& 4.124 & 4.586 & -9.500  & 0.000*** \\
Fairness& 4.180 & 4.551 & -7.235  & 0.000*** \\
Actionability& 3.444 & 4.052 & -9.356  & 0.000*** \\
Reasonableness& 3.764 & 4.216 & -8.015  & 0.000*** \\
Trustworthiness& 3.742 & 4.139 & -6.660  & 0.000*** \\
Satisfaction& 3.692 & 4.447 & -12.393 & 0.000*** \\
\bottomrule
\end{tabular}
\begin{tablenotes}
\footnotesize
\setlength{\baselineskip}{\normalbaselineskip}
\item Notes. Number of observations: 536. Ratings use 1 to 5 Likert scales where higher scores indicate better performance. Entries show item means by model, followed by two-sided \(t\)-tests. * $p < 0.05$, ** $p < 0.01$, *** $p < 0.001$.
\end{tablenotes}
\end{threeparttable}
\end{table}

}

\section{Conclusions and Discussion} \label{sec:conclusion}
This paper presents LEXMA, a systematic fine-tuning framework that combines reflection-augmented supervised fine-tuning and two stages of GRPO procedure to enhance explanation quality. The design keeps the decision conditioned on the explanation, uses GRPO with group comparisons to avoid training a separate reward model, and employs lightweight, rule-based rewards for readability and politeness, which together make the approach practical and cost efficient.
On the HMDA mortgage approval task, we empirically demonstrate that the fine-tuned Qwen3-4B attains competitive predictive accuracy while generating audience-appropriate narratives. Human evaluations from loan professionals and prospective applicants confirm the higher quality of the explanations generated by our fine-tuned model.

Our framework has several implications for marketing research and practice.
First, the LEXMA framework is general and extends beyond the specific context of mortgage decisions. Many marketing decisions combine algorithmic recommendation with communication to heterogeneous stakeholders, so any setting where AI assists decisions can benefit from explanations that are concise, faithful to case evidence, and tailored to different readers. One example is AI-driven personalization in marketing, where algorithms determine which offers, content, or prices to show on websites, apps, and email campaigns \citep{ma_machine_2020}. Marketing analytics teams and product managers need granular, model-centric explanations (e.g., which behavioral signals and segments drove a personalization decision), while end consumers may only see a succinct explanation embedded in an interface (e.g., ``you are seeing this offer because you frequently shop this category and recently browsed similar items''). LEXMA can support this multi-audience communication by using the correctness adapter to preserve the targeting policy and a tone adapter to generate expert rationales and consumer-friendly disclosures.

Second, the three-phase generation schema enables explanation quality to emerge from optimizing the prediction itself, since correctness-linked rewards flow through the explanation tokens that support accurate decisions. For marketing analytics and AI-enabled decision support, this provides a design pattern in which explanation is not a post hoc justification but a component of the optimization problem.

Third, our use of readability and politeness serves as a demonstration of a tone-oriented reward. In domain deployments, the tone-oriented reward can incorporate additional style and compliance objectives using rule-based calculation, supervised classifiers, or LLM-as-judge rubrics \citep{zheng_judging_2023}. In marketing applications, these objectives can be extended to enforce brand voice, avoid prohibited claims, or satisfy regulator-specified language for disclosures and consent.

Several limitations warrant consideration. First, our empirical analysis relies on the public HMDA Loan Application Register, which omits many credit-bureau and repayment variables for privacy reasons. Key measures such as detailed credit histories and granular delinquency patterns are therefore unavailable. Access to richer institution-level data would likely further enhance predictive performance and the faithfulness of explanations. Nonetheless, our findings indicate that even with a restricted feature set, the proposed framework can improve predictive power and explanation quality relative to strong baselines. Second, the expert evaluation draws on assessments from three seasoned practitioners and thirty decision pairs, and the consumer study relies on an online panel. These evaluations provide clear directional evidence but do not substitute for large-scale field deployment in live decision environments.
Finally, future research can further strengthen the faithfulness and robustness of explanations by augmenting group-relative optimization with richer reward composition and, where budget allows, incorporating a small sample of human annotations in domain-specific settings.

{\OneAndAHalfSpacedXI
\bibliographystyle{informs2014}
\bibliography{ref}
}
\newpage

\begin{APPENDICES}
\section{Additional Prompts}
For reflection-augmented supervised fine-tuning, we use the following prompts to generate ``reflect-then-predict'' responses.

{ \OneAndAHalfSpacedXI
\begin{tcolorbox}[
    enhanced,
    colframe=black,
    colback=gray!10,
    fonttitle=\bfseries,
    sharp corners=all,
    title=The Prompt Template for Loan Decision Reflection and Re-evaluation (Expert),
    width=\textwidth,
    boxrule=1pt,
    left=5pt,
    right=5pt,
    top=5pt,
    bottom=5pt,
    breakable
]
\textbf{System Role:}

You are a skilled loan expert that reflects on past incorrect predictions and re-does loan evaluations. You will be given a previous loan decision and explanation. You were unsuccessful in this task because you gave the wrong decision. First, diagnose a possible reason for failure, then re-evaluate the loan application. Specifically, you are tasked with evaluating the loan application and determining whether to approve or deny the loan. Provide a concise, expert-facing justification for the decision. Always respond strictly in the requested JSON format.

\textbf{User Role:}

Here is the previous trial loan information: \{summary\}

(END OF PREVIOUS LOAN INFORMATION)

Here are the previous explanation and decision: \{previous\_response\}

(END OF PREVIOUS EXPLANATION AND DECISION)

First, diagnose the possible reason for the incorrect decision. Then, re-evaluate the loan application and provide a new decision.

IMPORTANT: The Officer\_Message should be written as if this is the initial evaluation - present the decision and justification naturally without any indication that this is a re-evaluation or correction of a previous decision.

Respond only with JSON in this exact format: \texttt{\{\{"Prior\_failure\_analysis": "<Brief analysis of why the previous decision was incorrect>", "Officer\_Message": "<A short paragraph for the loan officer explaining and justifying the decision>", "Decision": "<Approved or Denied>"\}\}}.

Now start!
\end{tcolorbox}
}

{ \OneAndAHalfSpacedXI
\begin{tcolorbox}[
    enhanced,
    colframe=black,
    colback=gray!10,
    fonttitle=\bfseries,
    sharp corners=all,
    title=The Prompt Template for Loan Decision Reflection and Re-evaluation (Consumer),
    width=\textwidth,
    boxrule=1pt,
    left=5pt,
    right=5pt,
    top=5pt,
    bottom=5pt,
    breakable
]
\textbf{System Role:}

You are a skilled loan expert that reflects on past incorrect predictions and re-does loan evaluations. You will be given a previous loan decision and explanation. You were unsuccessful in this task because you gave the wrong decision. First, diagnose a possible reason for failure, then re-evaluate the loan application. Specifically, you are tasked with communicating directly with the applicant. Write in a polite, encouraging tone with clear, jargon-free language. Keep explanations actionable and easy to understand. If approving: summarize what went well and offer 1–2 simple tips for maintaining strong credit. If denying: list main reasons in plain language and give some specific, doable steps the applicant can take to improve their chances next time.

\textbf{User Role:}

Here is the previous trial loan information: \{summary\}

(END OF PREVIOUS LOAN INFORMATION)

Here are the previous explanation and decision: \{previous\_response\}

(END OF PREVIOUS EXPLANATION AND DECISION)

First, diagnose the possible reason for the incorrect decision. Then, re-evaluate the loan application and provide a new decision.

IMPORTANT: The Applicant\_Message should be written as if this is the initial evaluation - present the decision and guidance naturally without any indication that this is a re-evaluation or correction of a previous decision.

Respond only with JSON in this exact format: \texttt{\{\{"Prior\_failure\_analysis": "<Brief analysis of why the previous decision was incorrect>", "Applicant\_Message": "<Your short note to the applicant with the guidance above>", "Decision": "<Approved or Denied>"\}\}}.

Now start!
\end{tcolorbox}
}

\end{APPENDICES}

\end{document}